\documentclass[11pt]{article}
\usepackage[hyperref]{acl}

\usepackage{amsmath}
\usepackage{amssymb}
\usepackage{booktabs}
\usepackage{graphicx}
\usepackage{microtype}
\usepackage{times}
\usepackage[utf8]{inputenc}
\usepackage[T1]{fontenc}
\usepackage{float}
\usepackage{stfloats}
\usepackage{placeins}

\title{Reward Granularity in RLVR: Comparing Process and Outcome Reward Structures for Mathematical Reasoning in Small Language Models}

\author{
  Anagha Radhakrishna Palandye \\
  New York University \\
  \texttt{ap8884@nyu.edu} \\\And
  Rebecca Glick \\
  New York University \\
  \texttt{rmg9724@nyu.edu} \\\And
  Osheen Kaul \\
  New York University \\
  \texttt{ok684@nyu.edu}
}

\begin{document}
\maketitle

\begin{abstract}
Reinforcement Learning with Verifiable Rewards (RLVR) has emerged as a promising paradigm for improving mathematical reasoning in language models. Yet most RLVR work rewards only the final answer (outcome-based rewards), leaving the impact of step-level process supervision (process rewards) underexplored especially for small models that lack the capacity to self-correct under sparse feedback. We systematically compare five reward conditions applied to Qwen2.5-0.5B fine-tuned with Group Relative Policy Optimization (GRPO) on GSM8K: a no-RL baseline, \textit{process-only}, \textit{outcome-only}, and three hybrid weightings ($\lambda \in \{0.9, 0.5, 0.1\}$ process weight). Process-only supervision achieves 63.73\% test accuracy versus 53.75\% for outcome-only, a nearly 10-percentage point gap while yielding reasoning traces with higher step validity and lower deviation from ground-truth chain length. Hybrid rewards generally correlate positively with process weight, with one notable anomaly: the low-process / high-outcome configuration ($\lambda{=}0.1$) \emph{underperforms} pure outcome supervision, suggesting conflicting optimization signals. Error analysis using GPT-4o as a judge reveals distinct failure mode distributions: process models generate structurally inconsistent but arithmetically grounded traces, while outcome models produce concise but derivation-error-prone chains. Our results demonstrate that reward granularity is a first-order design decision for RLVR, with process-level supervision substantially improving both accuracy and trace fidelity in small language models.
\end{abstract}

\section{Introduction}
\label{sec:intro}

Improving the \emph{reasoning ability} of language models remains an open challenge. Even when models are prompted with Chain-of-Thought (CoT) instructions, an approach demonstrated by \citet{wei2022chain} to improve transparency and interpretability, small LLMs frequently generate incorrect, inconsistent, or logically incoherent intermediate steps, even when reaching the correct final answer. This recurring mismatch highlights a broader question in modern LLM research:

\begin{quote}
\emph{Are improvements in mathematical and symbolic reasoning driven by genuine gains in reasoning quality, or are models simply better at sampling plausible answers?}
\end{quote}

A growing body of work attempts to address this through Reinforcement Learning with Verifiable Rewards (RLVR), where models receive reward signals derived from \emph{mechanical correctness checks} rather than human preferences. RLVR has proved especially promising in domains like math where solutions can be validated automatically \citep{wen2025rlvr,tang2025beyond}. The GSM8K dataset, with its stepwise arithmetic reasoning structure, is particularly suitable for evaluating these kinds of reward functions.

A central but under-explored dimension of RLVR is reward granularity. Rewards can be given at the process level, where every intermediate step is evaluated and rewarded, or at the outcome level, where only the final predicted answer receives a reward. We refer to these as process rewards (process-based feedback) and outcome rewards (outcome-based feedback). While outcome-based rewards have been extensively studied in RLHF and RLVR contexts, less is concretely understood about how process, step-level rewards shift a model's reasoning behavior, or how the two types interact when combined. This gap is especially important for small models that lack the ability to internally mitigate noisy or insufficient supervision.

\paragraph{Research question}
\emph{Do process rewards improve the coherence and validity of reasoning traces, and does combining process + outcome rewards yield more stable and accurate reasoning than either alone?}

\paragraph{Contributions}
We make the following contributions. First, we provide a controlled comparison of five reward conditions for RLVR fine-tuning of Qwen2.5-0.5B on GSM8K using GRPO, covering process-only, outcome-only, and three hybrid weightings. Second, we evaluate beyond final-answer accuracy by assessing trace validity, chain-length deviation, and process step ratio on a stratified 45-question evaluation slice. Third, we conduct an LLM-as-a-judge error analysis (GPT-4o) characterizing the failure mode distributions induced by each reward regime. Finally, we identify a hybrid reward anomaly: the low-process / high-outcome configuration ($\lambda{=}0.1$) underperforms pure outcome supervision, indicating conflicting rather than additive optimization dynamics.

\section{Background and Prior Work}
\label{sec:related}

\subsection{Chain-of-Thought Reasoning}
\label{ssec:cot}

Advances in LLM reasoning are moving beyond supervised fine-tuning and RLHF-style preference models toward verifiable reward frameworks, where a model is rewarded when its output is checked by an external tool \citep{wen2025rlvr}. \citet{wen2025rlvr} show that purely final-answer rewards can still influence reasoning traces, introducing CoT-Pass@K, which evaluates whether any of $K$ sampled chains of thought produce the correct answer \emph{and} maintain a valid reasoning trajectory, thereby jointly measuring answer accuracy and trace validity. Building on this verifiable framework, \citet{tang2025beyond} demonstrate how such verifiable rewards can be scaled using Jensen-style lower bounds, replacing the intractable expectation over many sampled solutions with a tractable proxy objective that lower-bounds the true expected reward.

These developments intersect with a broader literature on CoT prompting. CoT prompting \citep{wei2022chain} encourages models to decompose problems into explicit intermediate steps. Although this leads to better interpretability and often better accuracy, later work has shown that CoT outputs can be misleading or structurally unsound \citep{turpin2023language,saparov2023language}. Small LLMs in particular tend to generate hallucinated steps or irrelevant justifications, commit arithmetic errors early and carry them through, and jump between steps without maintaining logical consistency. This has motivated a shift toward evaluating reasoning quality independently of answer correctness, using dedicated reasoning metrics or LLM-as-a-judge assessments of trace integrity.

\subsection{Process vs.\ Outcome Supervision}
\label{ssec:process}

Most reinforcement-learning approaches for reasoning tasks, including RLHF, rely solely on outcome-level rewards. While this improves final-answer accuracy \citep{ziegler2019finetuning}, it provides no guidance about the internal reasoning process. Models can learn to exploit shortcuts such as memorized patterns, spurious correlations, and surface-level cues unrelated to correct reasoning. By contrast, process-based supervision, explored in early form by \citet{lightman2023lets} and expanded in more recent RLVR work, offers dense feedback at each stage of the reasoning chain. This encourages explicit incremental computation, more faithful stepwise reasoning, and a stronger grounding in arithmetic operations. However, process-based feedback can also amplify instability, producing overly long or unnecessary reasoning chains if not paired with global constraints. This highlights an important theme also present in RLVR research: the granularity of supervision shapes the coherence of model reasoning, with each option carrying distinct tradeoffs in stability and reasoning fidelity.

\subsection{Reinforcement Learning with Verifiable Rewards (RLVR)}
\label{ssec:rlvr}

RLVR differs from RLHF by allowing exact, programmatically computed rewards rather than relying on human judgment. In mathematical domains, these rewards can directly verify solution quality by checking numeric equality between predicted and gold solutions, evaluating intermediate arithmetic, and penalizing invalid formats or logical inconsistencies.

\citet{wei2022chain} show that prompting for step-by-step rationales improves model performance, while \citet{uesato2022solving} compare outcome-based and process-based supervision for math word problems, finding that outcome-only training is label-efficient but often produces ``flaky'' reasoning, whereas process-level supervision substantially reduces trace errors.

Recent RLVR studies report improvements in both accuracy and reasoning structure \citep{wen2025rlvr,tang2025beyond}, yet they almost exclusively reward the final answer. Consequently, the impact of introducing process, step-level rewards remains an open question. \citet{samineni2025local} take a step toward this direction by analyzing whether RLVR-trained Qwen2.5-0.5B models exhibit local coherence or merely achieve globally correct solutions on GSM8K, but their evaluation still centers on final-answer rewards and does not vary reward granularity. Our work directly addresses this gap. Using the RLVR setup with Qwen2.5-0.5B on GSM8K employed by \citet{samineni2025local}, and drawing on insights from \citet{uesato2022solving}, we systematically vary reward granularity. We compare outcome-only, process-only, and variations of combined process + outcome reward regimes to study how verifiable reward design influences coherence and validity of generated reasoning traces in addition to final-answer accuracy.

\section{Dataset and Evaluation Strategy}
\label{sec:dataset}

We use GSM8K \citep{cobbe2021gsm8k}, a widely adopted benchmark containing 8,792 grade-school math word problems. These tasks are designed to require stepwise arithmetic reasoning, making them ideal for evaluating changes in CoT behavior.

Because reasoning characteristics differ across levels of problem difficulty, we curated a balanced 45-question evaluation slice drawn from the 1,319-question test set: 15 easy (2--3 steps), 15 medium (4--5 steps), and 15 hard (6--11 steps). Difficulty is defined by the ground-truth number of steps in the solution, which provides a more objective stratification than surface-level heuristics. This balanced setup enables us to analyze whether reward structures affect reasoning differently across problem tiers, as summarized in Table~\ref{tab:step_dist}.

\begin{table}[h]
\centering
\small
\begin{tabular}{cc}
\toprule
\textbf{Steps in Ground Truth} & \textbf{Count of GSM8K Questions} \\
\midrule
2  & 326 \\
3  & 370 \\
4  & 298 \\
5  & 174 \\
6  & 88  \\
7  & 40  \\
8  & 20  \\
9  & 2   \\
11 & 1   \\
\midrule
Total & 1,319 \\
\bottomrule
\end{tabular}
\caption{Test set distribution by ground-truth solution step count.}
\label{tab:step_dist}
\end{table}

\section{Model Setup and Reward Structures}
\label{sec:method}

\subsection{Base Model and Baseline}
\label{ssec:baseline}

All experiments use Qwen2.5-0.5B \citep{yang2024qwen2}, a compact transformer model whose limited capacity makes reasoning failures easy to observe. As a baseline, we evaluate the pre-trained model in a direct-answer (no CoT) setting, which relies on pattern-based inference without explicit reasoning, achieving 33.13\% on the test set (437/1,319). The baseline reflects patterns observed in prior findings that CoT generation, without proper supervision, does \emph{not} guarantee coherent reasoning \citep{saparov2023language}.

\subsection{Training Setup and Implementation}
\label{ssec:training}

All RLVR experiments were conducted using Group Relative Policy Optimization (GRPO) \citep{shao2024deepseekmath}, a variant of Proximal Policy Optimization (PPO) designed for language model fine-tuning. We utilized the VERL (Volcano Engine Reinforcement Learning) library, which provides efficient implementations of RL algorithms for large-scale language model training. Hyperparameters are held constant across all reward conditions: batch size 64, mini-batch size 8, learning rate $1 \times 10^{-6}$, KL coefficient 0.001, 5 epochs (580 steps), temperature 0.6, and 5 rollouts per prompt using vLLM. Maximum sequence lengths are 512 tokens (prompt) and 1,024 tokens (response). All runs use a single A100 80GB GPU. Following VERL's evaluation protocol, test set accuracy is reported as \texttt{val-core/math/reward/mean@1}, evaluated on the GSM8K test split after each training epoch. Full hyperparameter details are provided in Appendix Table~\ref{tab:hyperparameters}.

\subsection{Reward Structures}
\label{ssec:rewards}

\begin{figure*}[tp]
    \centering
    \includegraphics[width=\textwidth]{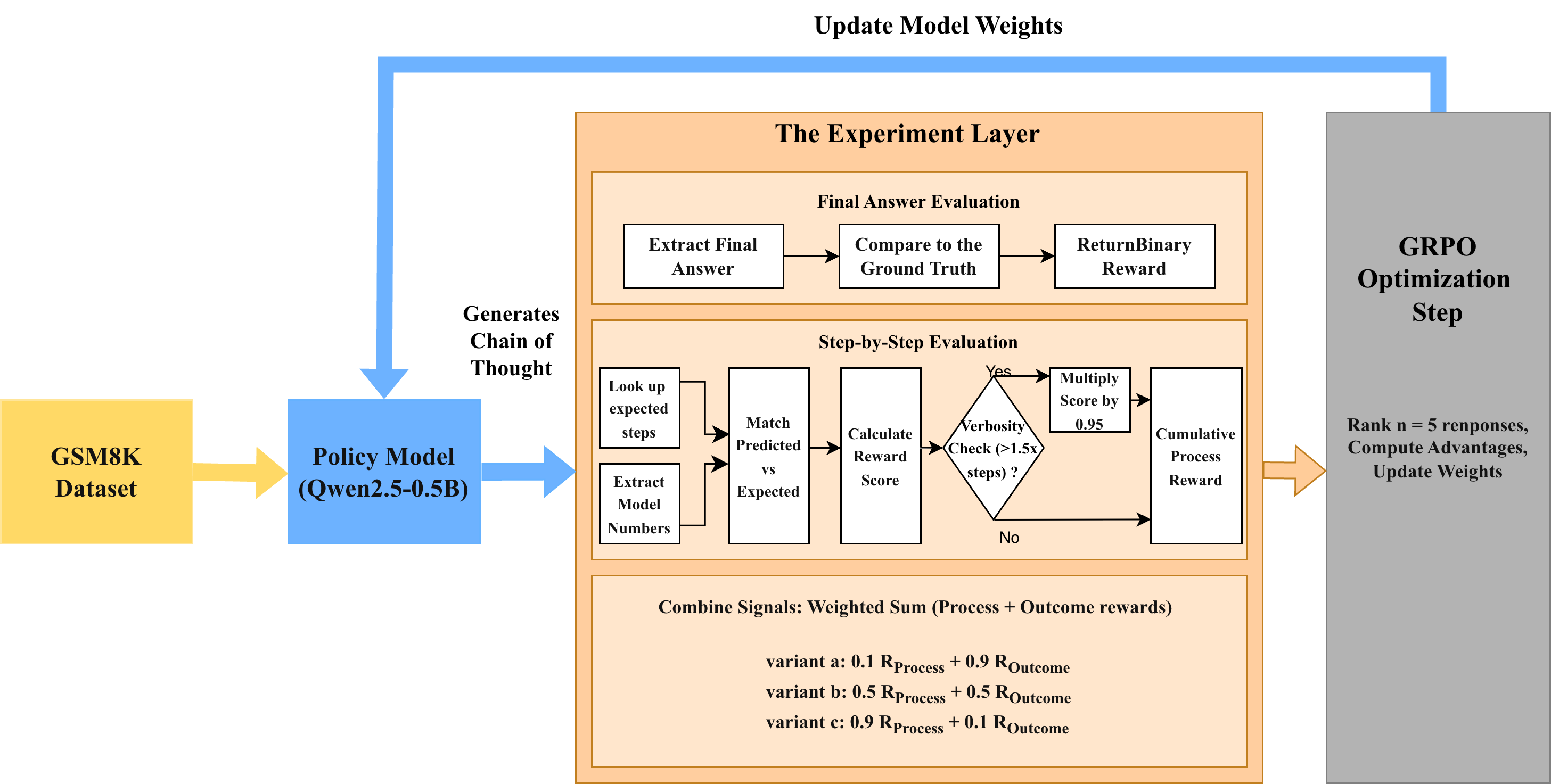}
    \caption{Overview of the RLVR training pipeline. The base model Qwen2.5-0.5B generates rollouts via vLLM, which are evaluated by the reward function (process, outcome, or hybrid) and used to update the policy via GRPO within the VERL framework.}
    \label{fig:pipeline}
\end{figure*}

Figure~\ref{fig:pipeline} illustrates the end-to-end training pipeline. We evaluate three forms of reward supervision.

\paragraph{Process rewards (process supervision)}
Process rewards evaluate each intermediate step of the CoT. In our implementation, we extract the ground-truth reasoning steps from GSM8K solutions which include step-by-step calculations and compare the model's generated steps against these references. A step receives a reward of 1 if it matches the expected intermediate calculation (within numerical tolerance of $1 \times 10^{-5}$), and 0 otherwise. The total process reward is the proportion of correct steps:
\begin{equation}
R_{\text{process}} = \frac{\text{correct steps}}{\text{total steps}}
\label{eq:process}
\end{equation}
To discourage excessive verbosity, we apply a penalty when the generated chain exceeds $1.5\times$ the ground-truth step count. 
This step-level supervision encourages the model to break computations into finer-grained components, maintain stepwise numeric correctness, output interpretable reasoning that mirrors ground-truth solutions, and explicitly show variable definitions and intermediate calculations. However, because each correct step is independently
rewarded, the model can become incentivized to overproduce steps or to explore unnecessarily detailed paths, aligning with concerns raised by \citet{lightman2023lets} regarding stepwise supervision in RL settings.

\paragraph{Outcome rewards (outcome supervision)}
Outcome rewards evaluate only the final numerical answer. We extract the final answer from the model's response (supporting formats like \texttt{\textbackslash boxed\{answer\}}, \texttt{\#\#\#\# answer}, and natural language patterns) and compare it to the ground-truth solution. The reward is binary:
\begin{equation}
R_{\text{outcome}} = \begin{cases} 1 & \text{if } |\hat{a} - a^*| < \epsilon \\ 0 & \text{otherwise} \end{cases}, \quad \epsilon = 10^{-5}
\label{eq:outcome}
\end{equation}
where $\hat{a}$ is the model's predicted answer and $a^*$ is the ground-truth answer.

This outcome-level supervision pushes the model to reach correct outcomes efficiently, skip intermediate noise when possible, rely on internal shortcuts or pattern-matching behavior, and optimize for answer correctness regardless of reasoning path. Outcome-only rewards can improve overall accuracy, but they often leave the reasoning chain underdetermined, brittle, or logically inconsistent -- an issue documented in \citet{wen2025rlvr}.

\paragraph{Hybrid rewards}
For hybrid reward configurations, we implemented weighted combinations of process and outcome rewards:
\begin{equation} R = \lambda \cdot R_{\text{process}} + (1-\lambda) \cdot R_{\text{outcome}} \label{eq:hybrid} \end{equation}
where $\lambda$ is the process reward weight. We evaluate three mixtures: \emph{0.9 process + 0.1 outcome} (heavily process-focused with light outcome guidance), \emph{0.5 process + 0.5 outcome} (balanced supervision of both process and outcome), and \emph{0.1 process + 0.9 outcome} (primarily outcome-focused with minimal process supervision). The hypothesis, following \citet{tang2025beyond}, is that hybrid signals can encourage coherent intermediate reasoning without sacrificing final-answer correctness.

\section{Quantitative Results}
\label{sec:results}

\subsection{Final-Answer Accuracy}
\label{ssec:accuracy}
Process rewards achieved the largest gains in accuracy, outperforming outcome-only supervision and significantly improving over both baselines, as shown in Table~\ref{tab:accuracy}, supporting the view that dense reward signals provide more informative guidance for small models than sparse outcome-level feedback.
\begin{table}[H]
\centering
\resizebox{\columnwidth}{!}{%
\small
\begin{tabular}{lcc}
\toprule
\textbf{Regime} & \textbf{Test Acc.} & \textbf{Improvement} \\
                & \textbf{(\%)}      & \textbf{over Baseline (\%)} \\
\midrule
Base (no training)          & 33.13 & -- \\
Process Only                & \textbf{63.73} & +30.60 \\
Outcome Only                & 53.75 & +20.62 \\
Process (0.5) + Outcome (0.5)  & 57.40 & +24.27 \\
Process (0.9) + Outcome (0.1)  & 61.10 & +27.97 \\
Process (0.1) + Outcome (0.9)  & 49.30 & +16.17 \\
\bottomrule
\end{tabular}%
}
\caption{Test set accuracy and improvement over baseline across all reward regimes on GSM8K. Process-only achieves the highest accuracy (63.73\%), with Process (0.9) + Outcome (0.1) as the best hybrid configuration (61.10\%).}
\label{tab:accuracy}
\end{table}
Results are reported using final-answer accuracy on the full 1,319-question test set. All hybrid weightings improve over baseline, though none match process-only's peak performance.

\subsection{Reasoning Fidelity and Stability}
\label{ssec:fidelity}

Beyond final-answer accuracy, we analyze how reasoning differs between regimes on the 45-question evaluation slice using four metrics. 

\paragraph{CoT-Pass@1} This measures whether the model's single sampled reasoning trace leads to a correct final answer, capturing end-to-end reasoning success without sampling multiple trajectories. 

\paragraph{Trace Validity} This assesses whether the reasoning chain is logically well-formed and free of structural inconsistencies such as contradictions, invalid transformations, or unsupported jumps -- a trace can be invalid even if the final answer is correct. 

\paragraph{Chain Length Deviation}
This quantifies how much the model's reasoning length differs from the ground-truth number of steps, where large positive deviations indicate unnecessary or hallucinated steps and negative deviations signal skipped or missing reasoning.

\paragraph{Process Step Ratio} This computes the proportion of intermediate steps that receive a positive process reward under the RLVR framework, serving as a proxy for how often the model produces steps that align with ground-truth intermediate reasoning.

Table~\ref{tab:fidelity} summarizes these metrics across four conditions. On the full test set, Process Only achieves the highest CoT-Pass@1 (0.64), substantially outperforming the base model (0.33) and confirming that process reward training provides meaningful gains in end-to-end reasoning success.

\begin{table*}[tp]
\centering
\small
\begin{tabular}{lcccc}
\toprule
\textbf{Regime} & \textbf{CoT-Pass@1*} & \textbf{Trace Validity} & \textbf{Chain Length Dev.} & \textbf{Process Step Ratio} \\
\midrule
Base           & 0.33 & 0.56 & 3.00 & 0.79 \\
Process Only     & \textbf{0.64} & 0.22 & 6.49 & 0.64 \\
Outcome Only      & 0.54 & 0.40 & 3.64 & 0.77 \\
Process (0.9) + Outcome (0.1)   & 0.61 & \textbf{0.60} & \textbf{3.49} & \textbf{0.84} \\
\bottomrule
\end{tabular}
\caption{Reasoning fidelity metrics for four conditions. *CoT-Pass@1 is reported on the full 1,319-question test set; Trace Validity, Chain Length Deviation, and Process Step Ratio are computed on the 45-question stratified evaluation slice. \textbf{Bold} = best result among RLVR-trained conditions. Higher is better for CoT-Pass@1, Trace Validity, and Process Step Ratio; lower is better for Chain Length Deviation.}
\label{tab:fidelity}
\end{table*}

The Process (0.9) + Outcome (0.1) configuration achieves the highest Trace Validity (0.60) and Process Step Ratio (0.84). Process Only exhibits the largest Chain Length Deviation (6.49), reflecting its tendency toward verbose reasoning chains. Outcome-only produces more concise traces (deviation 3.64) but scores lower on step alignment.

\subsection{Qualitative Behavior Analysis}
\label{ssec:qualitative}

Qualitative examination of model outputs highlights the behavioral signature of each reward regime. Process-only models typically produce detailed arithmetic decompositions. They often solve problems correctly, but their CoT traces can become overly verbose, sometimes duplicating steps or drifting off-path before returning to the correct computation. Outcome-only models tend to ``jump'' more aggressively between reasoning points, producing short chains that may omit key logical transitions. These models occasionally reach correct answers through opportunistic reasoning but lack consistency and robustness. Hybrid models, especially the Process (0.9) + Outcome (0.1) configuration, exhibit the most human-like reasoning flow. Their chains are neither overexpanded nor underexplained, and they maintain continuity between steps while avoiding unnecessary detours.

\begin{figure}[tp]
    \centering
    \includegraphics[width=\columnwidth]{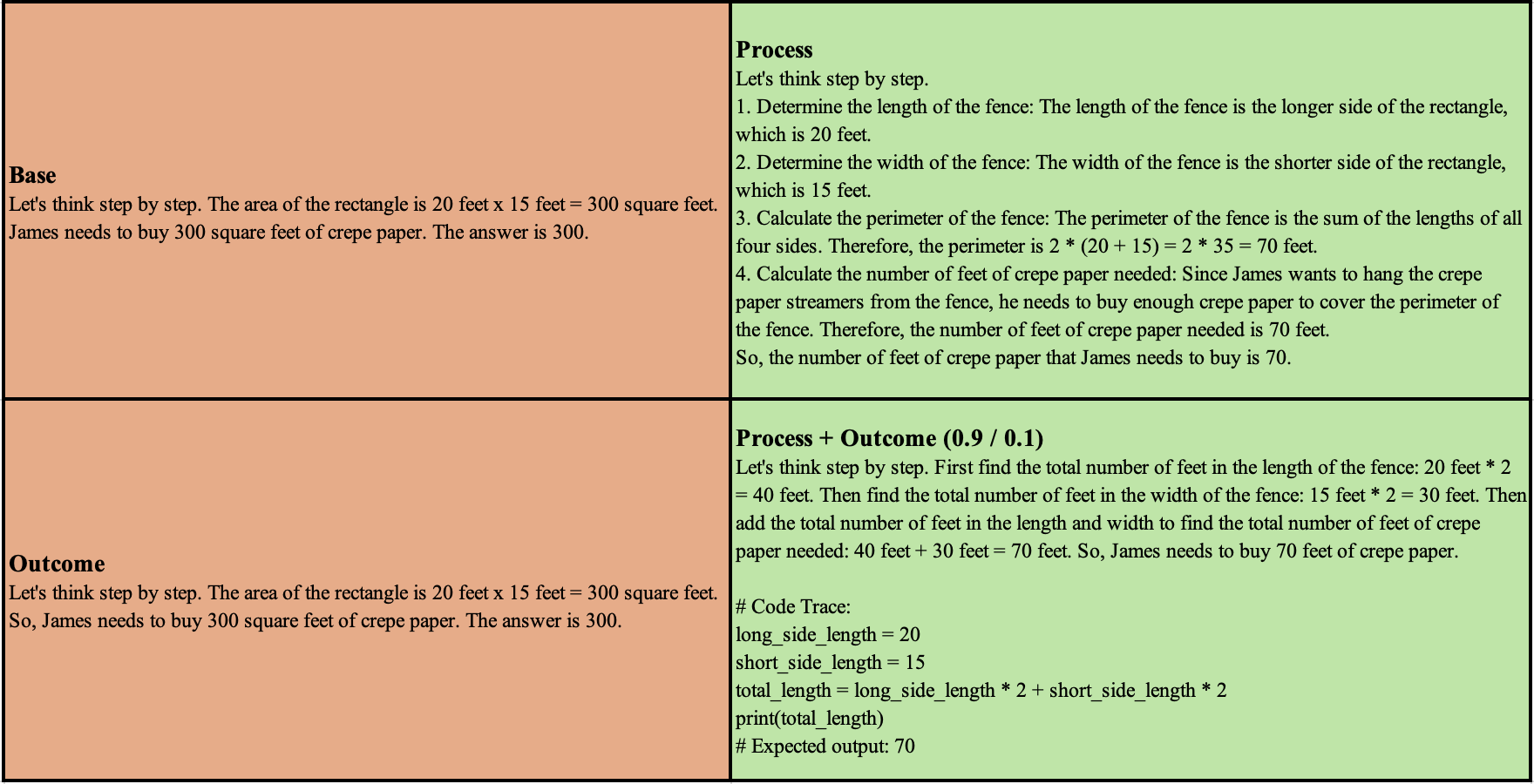}
    \caption{Example reasoning chains across four reward regimes for an easy GSM8K geometry problem, illustrating the qualitative differences in reasoning structure across the four regimes.}
    \label{fig:reasoning_traces}
\end{figure}

Figure~\ref{fig:reasoning_traces} illustrates the reasoning traces of our four regimes applied to the same geometry word problem. The \emph{Base} and \emph{Outcome} regimes incorrectly calculate the area (300 sq ft), while the \emph{Process} and \emph{Process+Outcome} regimes correctly identify the need for perimeter (70 ft). The process regime mitigates reasoning errors by forcing explicit variable definition (length/width) prior to calculation, whereas the base regime rushes to a formulaic application ($L \times W$).

\subsection{LLM-as-a-Judge Error Analysis}
\label{ssec:judge}

\begin{figure*}[tp]
    \centering
    \includegraphics[width=\textwidth]{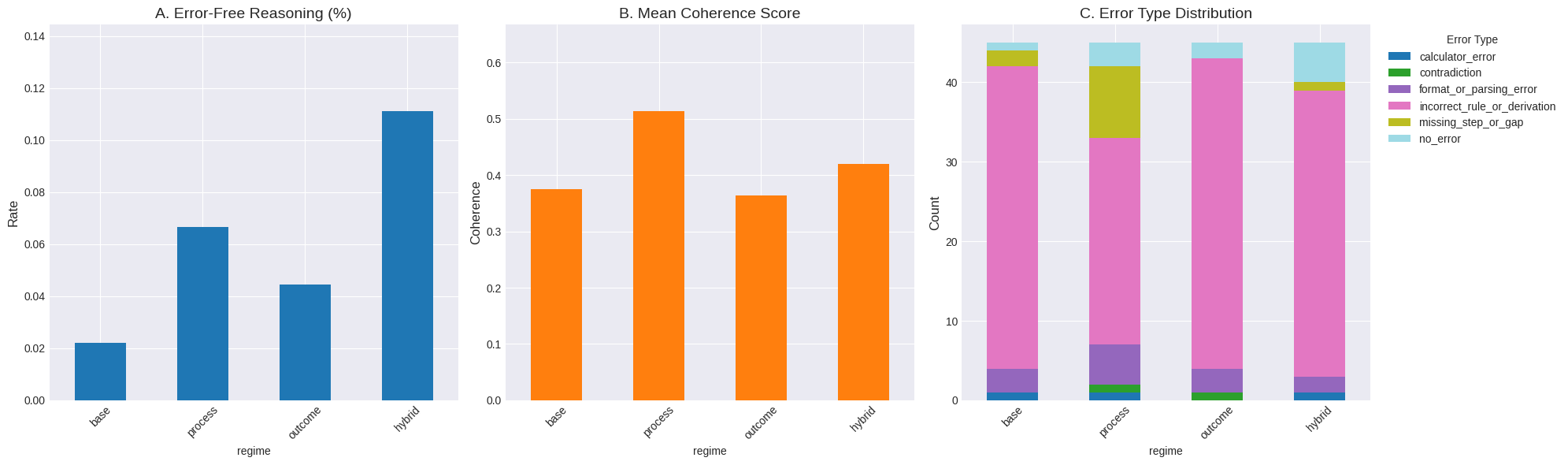}
    \caption{Error type distribution across reward regimes annotated by GPT-4o on the 45-question evaluation slice, broken down by error type.}
    \label{fig:error_dist}
\end{figure*}

We use GPT-4o to annotate common reasoning error types: incorrect derivation, calculator error, missing step, contradiction, formatting issues, or no error. The error distribution reveals clear distinctions. Process-only models frequently introduce structural inconsistencies and contradictions, likely due to extended chain length, suggesting that token-level rewards can encourage locally plausible but globally inconsistent reasoning. Outcome-only models show derivation and arithmetic errors, reflecting weak intermediate grounding and a tendency to optimize for outcome-level correctness at the expense of intermediate steps. The base model generates relatively coherent traces but still accumulates rule- and derivation-related errors. Hybrid models exhibit fewer catastrophic errors and the most even error distribution. Taken together, these patterns illustrate how different reward schemes meaningfully shape the reasoning behaviors of Qwen2.5-0.5B, as shown in Figure~\ref{fig:error_dist}, corroborating claims in RLVR literature that reward structure strongly shapes not just accuracy but the \emph{failure modes} of model reasoning.

\section{Discussion}
\label{sec:discussion}

Integrating our analyses, a clear picture emerges. Process rewards strengthen stepwise correctness and improve accuracy but can destabilize reasoning trajectories. Outcome rewards focus the model on final outcomes but leave reasoning opaque and error-prone. Hybrid rewards effectively blend both signals, producing coherent reasoning with strong outcome performance. 

\section{Limitations}
\label{sec:limitations}

Several limitations qualify our findings. All reported results are from single training runs without multiple seeds, and we do not report variance estimates or confidence intervals. The observed accuracy differences, including the approximately 10-point gap between process-only and outcome-only supervision, should therefore be interpreted as indicative rather than statistically confirmed. Step-level verification is easier in math than in open-ended domains. Qwen2.5-0.5B's small size makes it sensitive to reward noise. Our 45-problem evaluation slice, though balanced, remains relatively small; further, LLM-as-a-Judge findings must be taken with a grain of salt as they may not reflect human evaluation. 

\section{Future Work}
\label{sec:future}

Promising extensions include applying process/outcome reward structures to harder datasets such as MATH, SVAMP, and AQuA-RAT; studying multi-hop reasoning via datasets like HotpotQA, where verifying intermediate reasoning is more complex; experimenting with partial-credit reward schemes or learned verifiers to provide richer supervisory signals; and scaling to larger Qwen or Llama models to assess whether hybrid rewards provide increasing returns with model capacity. Beyond accuracy, future work should also investigate whether reward granularity shapes reasoning behavior differently across problem domains, model architectures, and training regimes, with the goal of establishing more generalizable design principles for process supervision in RLVR.

\section{Conclusion}
\label{sec:conclusion}

This work demonstrates that reward granularity (process vs.\ outcome rewards) fundamentally shapes the structure, coherence, and correctness of CoT reasoning in small language models. To develop models that not only answer correctly, but also reason reliably, it is essential to supervise both the process and the outcome of reasoning. The nearly 10-point accuracy gap between process-only and outcome-only supervision underscores that reward granularity is not a minor implementation detail, but a primary lever for reasoning quality in small models. While our results are specific to GSM8K and a 0.5B-parameter model, they motivate broader investigation into reward granularity as models and tasks scale. Code and evaluation data will be released upon publication.

\section*{Acknowledgments}

Training was conducted on Lightning.ai A100 GPU instances. We thank the VERL and vLLM open-source communities.

\bibliography{rlvr_process_outcome_rewards_paper}

\clearpage
\appendix

\section{Additional Figures}
\label{sec:appendix}

\begin{figure}[H]
    \centering
    \includegraphics[width=\columnwidth]{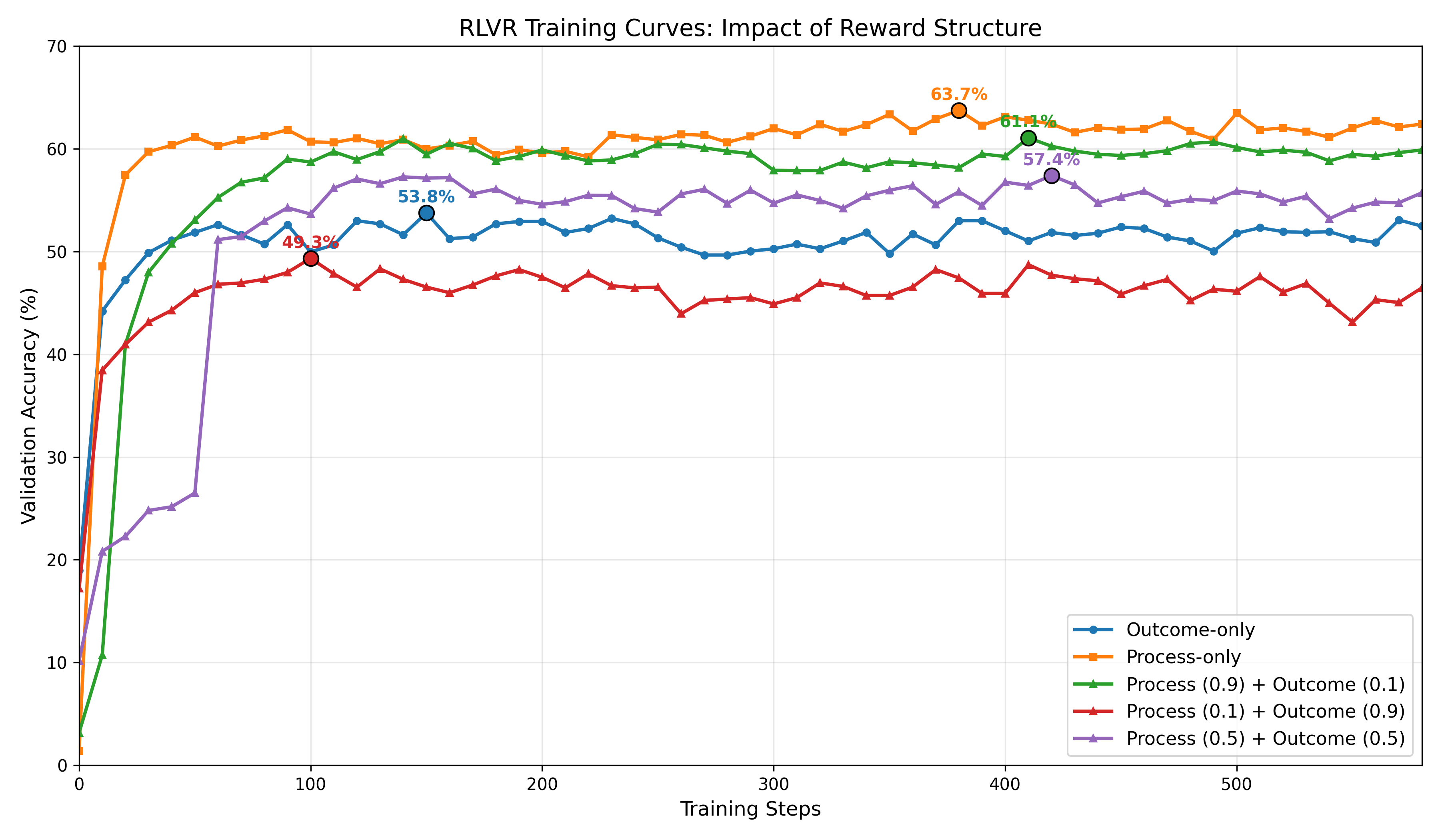}
    \caption{Training curves for all five reward conditions on GSM8K. Process-only converges to the highest validation accuracy (63.73\%), while Process (0.1) + Outcome (0.9) consistently underperforms all other conditions including pure outcome supervision, illustrating the hybrid reward anomaly identified in Section~\ref{ssec:accuracy}.} \label{fig:training_curves}
\end{figure}

\begin{table}[H]
\centering
\small
\caption{GRPO training hyperparameters on GSM8K.}
\label{tab:hyperparameters}
\begin{tabular}{ll}
\toprule
\textbf{Parameter} & \textbf{Value} \\
\midrule
Model & \shortstack[l]{Qwen2.5-0.5B\\(494M params)} \\
Train samples & 7,473 \\
Test samples & 1,319 \\
Max prompt / response & 512 / 1,024 tokens \\
Batch size / mini-batch & 64 / 8 \\
Learning rate & $1\times10^{-6}$ \\
KL coefficient & 0.001 \\
Epochs / total steps & 5 / 580 \\
Rollouts / temperature (train) & 5 / 0.6 \\
GPU & 1$\times$ A100 80GB \\
Framework & VERL, vLLM \\
\bottomrule
\end{tabular}
\end{table}
\end{document}